%% file: 00-main.tex
\documentclass[11pt,a4paper]{article}
\usepackage[hyperref]{acl2020}
\usepackage{times}
\usepackage{latexsym}

\usepackage[T5,T2A,T1]{fontenc}
\usepackage[utf8]{inputenc}

\usepackage{url}
\usepackage{nicefrac}
\usepackage{calc}
\usepackage{xcolor}
\usepackage{graphicx}
\usepackage{multirow}

\usepackage{amsthm}
\usepackage{amssymb}
\usepackage{amsmath}
\usepackage{amsfonts}
\usepackage[toc]{appendix}
\usepackage{color, colortbl}
\usepackage{adjustbox}
\usepackage{tabularx}
\usepackage{booktabs}
\usepackage{subcaption}
\usepackage{makecell}
\usepackage{arydshln}
\usepackage{tikz}
\usetikzlibrary{positioning}

\graphicspath{{img}}

\definecolor{Gray}{gray}{0.92}
\newcolumntype{Y}{>{\centering\arraybackslash}X}

\usepackage{microtype}

\aclfinalcopy 

\newcommand{\mytexttt}[1]{\texttt{\small #1}}

\title{\textit{Span-ConveRT}: Few-shot Span Extraction for Dialog with \\ Pretrained Conversational Representations}

\author{Sam Coope$^{\mathbf{1}}$\thanks{{ } Both authors contributed equally to the work. The work of TF was done during an internship at PolyAI.}, ~Tyler Farghly$^{\mathbf{2}*}$, Daniela Gerz$^{\mathbf{1}}$, Ivan Vuli\'{c}$^{\mathbf{1,3}}$, Matthew Henderson$^{\mathbf{1}}$\\
$^{\mathbf{1}}$  PolyAI Limited, London, UK\\
$^{\mathbf{2}}$ Imperial College London, UK \\
$^{\mathbf{3}}$ Language Technology Lab, University of Cambridge, UK\\
\texttt{sam@polyai.com}}

\date{}

\begin{document}
\maketitle
\begin{abstract}
We introduce \textit{Span-ConveRT}, a light-weight model for dialog slot-filling which frames the task as a turn-based span extraction task. This formulation allows for a simple integration of conversational knowledge coded in large pretrained conversational models such as ConveRT \cite{Henderson:2019convert}. We show that leveraging such knowledge in Span-ConveRT is especially useful for few-shot learning scenarios: we report consistent gains over \textbf{1)} a span extractor that trains representations from scratch in the target domain, and \textbf{2)} a BERT-based span extractor. In order to inspire more work on span extraction for the slot-filling task, we also release \textsc{restaurants-8k}, a 
new challenging data set of 8,198 utterances, compiled from actual conversations in the restaurant booking domain.








\end{abstract}

\section{Introduction}
\label{s:intro}
\input{01-intro.tex}

\section{Methodology: Span-ConveRT}
\label{s:methodology}
\input{02-methodology.tex}

\section{Experimental Setup}
\label{s:exp}
\input{03-experimental.tex}

\section{Results and Discussion}
\label{s:results}

\input{04-results.tex}


\section{Conclusion and Future Work}
\label{s:conclusion}
\input{06-conclusion.tex}

\section*{Acknowledgments}
We thank the three anonymous reviewers for their helpful suggestions and feedback. We are grateful to our colleagues at PolyAI, especially Georgios Spithourakis and I\~{n}igo Casanueva, for many fruitful discussions and suggestions.

\bibliography{refs, other_refs}
\bibliographystyle{acl_natbib}

\clearpage
\label{s:appendix}
\input{07-appendix.tex}

\end{document}

%% file: 01-intro.tex

Conversational agents are finding success in a wide range of well-defined tasks such as customer support, restaurant, train or flight bookings \cite{Hemphill:1990,Williams:2012b,ElAsri:2017sigdial,Budzianowski:2018emnlp}, language learning \cite{Raux:2003,Chen:2017survey}, and also in domains such as healthcare \cite{Laranjo:2018} or entertainment \cite{Fraser:2018iva}. Scaling conversational agents to support new domains and tasks, and particular system behaviors is a highly challenging and resource-intensive task: it critically relies on expert knowledge and domain-specific labeled data \cite{Williams:2014sigdial,Wen:17,Wen:2017icml,Liu:2018naacl,Zhao:2019naacl}.

Slot-filling is a crucial component of any task-oriented dialog system \cite{Young:02is,young:10b,Bellegarda:2014siri}. For instance, a conversational agent for restaurant bookings must fill all the slots \textit{date}, \textit{time} and \textit{number of guests} with correct values given by the user (e.g. \textit{tomorrow}, \textit{8pm}, \textit{3 people}) in order to proceed with a booking. A particular challenge is to deploy slot-filling systems in \textit{low-data regimes} (i.e., \textit{few-shot learning} setups), which is needed to enable quick and wide portability of conversational agents. Scarcity of in-domain data has typically been addressed using domain adaption from resource-rich domains, e.g. through multi-task learning \cite{Jaech:2016is,Goyal:2018naacl} or ensembling \cite{Jha:2018naacl,Kim:2019conll}.

In this work, we approach slot-filling as a \textit{turn-based span extraction} problem similar to \newcite{rastogi2019towards}: in our \textit{Span-ConveRT} model we do not restrict values to fixed categories, and simultaneously allow the model to be entirely independent of other components in the dialog system. In order to facilitate slot-filling in resource-lean settings, our main proposal is the effective use of knowledge coded in representations transferred from large general-purpose conversational pretraining models, e.g., the ConveRT model trained on a large Reddit data set \cite{Henderson:2019convert}. 

\begin{figure}[t!]
    \includegraphics[width=0.485\textwidth]{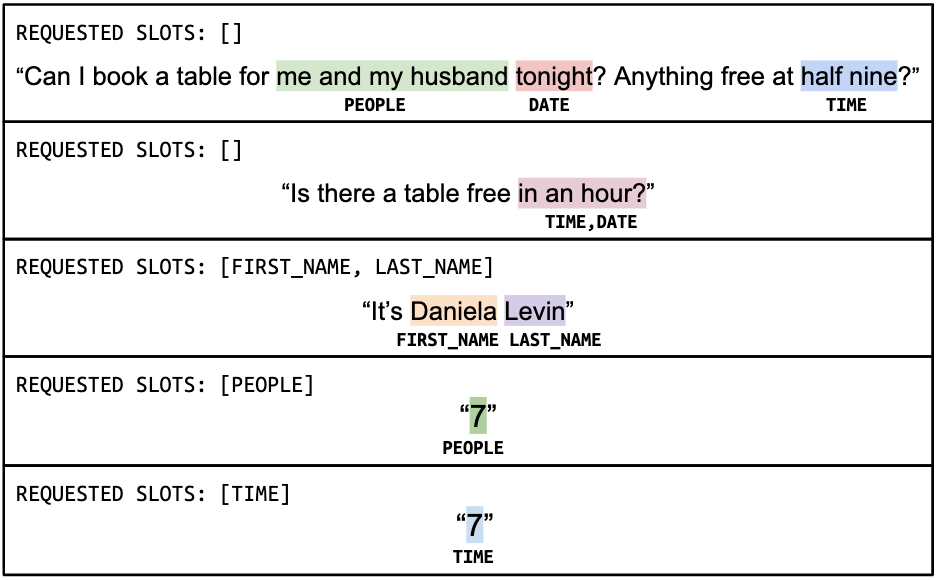} 
    \centering
    \caption{Turn-based span extraction with the new \textsc{restaurants-8k} data set. Note how the requested slot feature is needed to differentiate time or party size in short utterances like ``7''. The single-turn examples are extracted from different conversations.
    }
    \label{fig:exampledata}
    \vspace{-1mm}
\end{figure}

To help guide other work on span extraction-based slot-filling, we also present a new data set of 8,198 user utterances from a commercial restaurant booking system: \textsc{restaurants-8k}. The data set spans 5 slots (\textit{date}, \textit{time}, \textit{people}, \textit{first name}, \textit{last name}) and consists of actual user utterances collected ``in the wild''. This comes with a broad range of natural and colloquial expressions,\footnote{For instance, a value for the slot \textit{people} can either be a number like \textit{7}, or can be expressed fully in natural language, e.g., \textit{me and my husband}.} as illustrated in Figure~\ref{fig:exampledata}, which makes it both a natural and challenging benchmark. Each training example is a dialog turn annotated with the slots requested by the system and character-based span indexing for all occurring values. 

As our key findings show, conversational pretraining is instrumental to span extraction performance in few-shot setups. By using subword representations transferred from ConveRT \cite{Henderson:2019convert}, we demonstrate that: 1) our ConveRT-backed span extraction model outperforms the model based on transferred BERT representations, and 2) it also yields consistent gains over a span extraction model trained from scratch in the target domains, with large gains reported in few-shot scenarios. We verify both findings on the new \textsc{restaurants-8k} data set, as well as on four \textsc{dstc8}-based data sets \cite{rastogi2019towards}. All of the data sets used in this work are available online at: \url{https://github.com/PolyAI-LDN/task-specific-datasets}.

%% file: 02-methodology.tex
Before we delve into describing the core methodology, we note that in this work we are not concerned with the task of normalizing extracted spans to their actual values: this can be solved effectively with rule-based systems after the span extraction step for cases such as times, dates, and party sizes. There exist hierarchical rule-based parsing engines (e.g., Duckling) that allow for parsing times and dates such as \textit{``the day after next Tuesday''}. Further, phrases such as \textit{``Me and my wife and 2 kids''} can be parsed using singular noun and number counts in the span with high precision.


\vspace{1.6mm}
\noindent \textbf{Span Extraction for Dialog.}
We have recently witnessed increasing interest in \textit{intent-restricted} approaches \cite{coucke2018snips,Goo:18, chen} for slot-filling. In this line of work, slot-filling is treated as a span extraction problem where slots are defined to occur only with certain intents. This solves the issue of complex categorical modeling but makes slot-filling dependent on an intent detector. Therefore, we propose a framework that treats slot-filling as a fully \textit{intent-agnostic} span extraction problem. Instead of using rules to constrain the co-occurrence of slots and intents, we identify a slot as either a single span of text or entirely absent. This makes our approach more flexible than prior work; it is fully independent of other system components. Regardless, we can explicitly capture turn-by-turn context by adding an input feature denoting whether a slot was requested for this dialog turn (see Figure~\ref{fig:exampledata}).

\vspace{1.6mm}
\noindent \textbf{Pretrained Representations.}
Large-scale pretrained models have shown compelling benefits in a plethora of NLP applications \cite{Devlin:2018arxiv,Liu:2019roberta}: such models drastically lessen the amount of required task/domain-specific training data with in-domain fine-tuning. This is typically achieved by adding a task-specific output layer to a large pretrained encoder and then fine-tuning the entire model \cite{Xie:2019}. However, this process requires a fine-tuned model for each slot or domain, rather than a single model shared across all slots and domains. This adds a large memory and computational overhead and makes the approach impractical in real-life applications. Therefore, we propose to keep the pretrained encoder models fixed in order to emulate a production system where a single encoder model is used.\footnote{In other words, we do not fine-tune the parameters of the pretrained encoders which would require running a separate encoder for each slot. This would mean, for example, we would need 100 fine-tuned encoders running in production to support 100 different slots. As the encoder models have both high memory and runtime requirements, this would drastically increase the running costs of a conversational system.}



\vspace{1.6mm}
\noindent \textbf{Underlying Representation Model: ConveRT.} ConverRT \cite{Henderson:2019convert} is a light-weight sentence encoder implemented as a dual-encoder network that models the interaction between inputs/contexts and relevant (follow-up) responses. In other words, it performs conversational pretraining based on response selection on the Reddit corpus \cite{Henderson:2019convert, Henderson:2019acl}. It utilizes subword-level tokenization and is very compact and resource-efficient (i.e. it is 59MB in size and can be trained in less than 1 day on 12 GPUs) while achieving state-of-the-art performance on conversational tasks \cite{Casanueva:2020ws,Bunk:2020arxiv}.  Through pretrained ConveRT representations, we can leverage conversational cues from over 700M conversational turns for the few-shot span extraction task.\footnote{As we show later in \S\ref{s:results}, we can also leverage BERT-based representations in the same span extraction framework, but our ConveRT-based span extractors result in higher performance.}





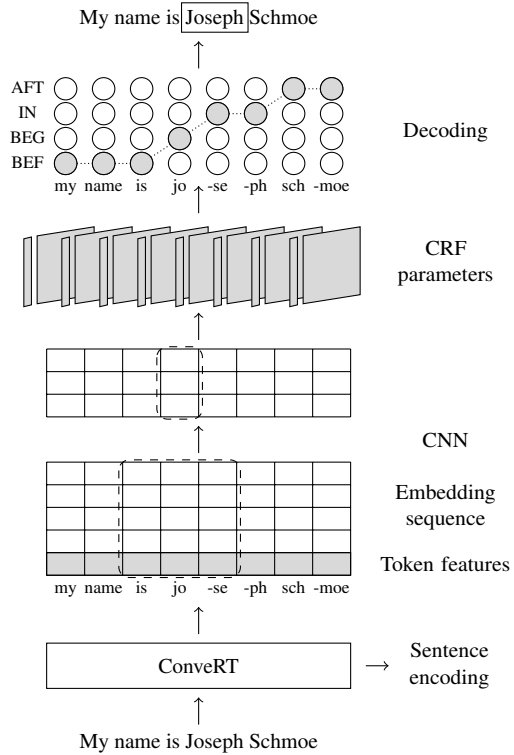
\begin{figure}
    \centering
\begin{tikzpicture}[xscale=0.5, yscale=0.3, every node/.style={scale=0.73}]

\draw (3.55, 25) rectangle (5.3, 26.3);
\draw [text centered] (4, 25.6) node {My name is  Joseph Schmoe};

\path [->] (4, 23.4) edge (4, 24.6);

\draw [text centered, text width=2.5cm] (10.5, 20.6) node {Decoding};

\path [densely dotted] (6.5, 22.5) edge (7.5, 22.5);
\path [densely dotted] (5.5, 21.4) edge (6.5, 22.5);
\path [densely dotted] (4.5, 21.4) edge (5.5, 21.4);
\path [densely dotted] (3.5, 20.3) edge (4.5, 21.4);
\path [densely dotted] (2.5, 19.2) edge (3.5, 20.3);
\path [densely dotted] (1.5, 19.2) edge (2.5, 19.2);
\path [densely dotted] (0.5, 19.2) edge (1.5, 19.2);
\foreach \x in {0,...,7} {
	\foreach \y in {0,...,3} {
		\draw [fill=white] (\x + 0.5, 19.2 + \y*1.1) ellipse (8.4pt and 14pt);
	}
}

\draw [fill=black!15] (0.5, 19.2) ellipse (8.4pt and 14pt);
\draw [fill=black!15] (1.5, 19.2) ellipse (8.4pt and 14pt);
\draw [fill=black!15] (2.5, 19.2) ellipse (8.4pt and 14pt);
\draw [fill=black!15] (3.5, 20.3) ellipse (8.4pt and 14pt);
\draw [fill=black!15] (4.5, 21.4) ellipse (8.4pt and 14pt);
\draw [fill=black!15] (5.5, 21.4) ellipse (8.4pt and 14pt);
\draw [fill=black!15] (6.5, 22.5) ellipse (8.4pt and 14pt);
\draw [fill=black!15] (7.5, 22.5) ellipse (8.4pt and 14pt);

\draw [font=\small, minimum height = 15pt, text height = 15pt, text depth = 15pt] (-0.5, 19.0) node {BEF};
\draw [font=\small, minimum height = 15pt, text height = 15pt, text depth = 15pt] (-0.5, 20.1) node {BEG};
\draw [font=\small, minimum height = 15pt, text height = 15pt, text depth = 15pt] (-0.5, 21.2) node {IN};
\draw [font=\small, minimum height = 15pt, text height = 15pt, text depth = 15pt] (-0.5, 22.3) node {AFT};

\draw [font=\small, minimum height = 15pt, text height = 15pt, text depth = 15pt] (0.5, 18) node {my};
\draw [font=\small, minimum height = 15pt, text height = 15pt, text depth = 15pt] (1.5, 18) node {name};
\draw [font=\small, minimum height = 15pt, text height = 15pt, text depth = 15pt] (2.5, 18) node {is};
\draw [font=\small, minimum height = 15pt, text height = 15pt, text depth = 15pt] (3.5, 18) node {jo};
\draw [font=\small, minimum height = 15pt, text height = 15pt, text depth = 15pt] (4.5, 18) node {-se};
\draw [font=\small, minimum height = 15pt, text height = 15pt, text depth = 15pt] (5.5, 18) node {-ph};
\draw [font=\small, minimum height = 15pt, text height = 15pt, text depth = 15pt] (6.5, 18) node {sch};
\draw [font=\small, minimum height = 15pt, text height = 15pt, text depth = 15pt] (7.5, 18) node {-moe};

\path [->] (4, 17) edge (4, 18.2);

\draw [text centered, text width=2.5cm] (10.5, 14.8) node {CRF \\parameters};
\foreach \x in {0,...,7} {
	\draw [fill=black!15] (\x - 0.25, 16) -- (\x + 1.25, 16.4) -- (\x + 1.25, 13.4) -- (\x - 0.25, 13) -- cycle;
	
	\draw [fill=black!15] (\x - 0.6, 15.87) -- (\x - 0.4, 15.95) -- (\x - 0.4, 12.95) -- (\x - 0.6, 12.87) -- cycle;

}

\path [->] (4, 11.4) edge (4, 12.6);

\draw (0, 8) grid (8, 11);
\draw[dashed, rounded corners] (2.9, 11.1) rectangle (4.1, 7.9);

\draw [text centered, text width=2.5cm] (10.5, 7) node {CNN};
\path [->] (4, 6.4) edge (4, 7.6);

\draw [text centered, text width=2.5cm] (10.5, 4) node {Embedding sequence};
\draw [text centered, text width=2.5cm] (10.5, 1.5) node {Token features};
\draw [fill=black!15] (0, 1) rectangle (8, 2);
\draw (0, 1) grid (8, 6);

\draw [font=\small, minimum height = 15pt, text height = 15pt, text depth = 15pt] (0.5, 0.2) node {my};
\draw [font=\small, minimum height = 15pt, text height = 15pt, text depth = 15pt] (1.5, 0.2) node {name};
\draw [font=\small, minimum height = 15pt, text height = 15pt, text depth = 15pt] (2.5, 0.2) node {is};
\draw [font=\small, minimum height = 15pt, text height = 15pt, text depth = 15pt] (3.5, 0.2) node {jo};
\draw [font=\small, minimum height = 15pt, text height = 15pt, text depth = 15pt] (4.5, 0.2) node {-se};
\draw [font=\small, minimum height = 15pt, text height = 15pt, text depth = 15pt] (5.5, 0.2) node {-ph};
\draw [font=\small, minimum height = 15pt, text height = 15pt, text depth = 15pt] (6.5, 0.2) node {sch};
\draw [font=\small, minimum height = 15pt, text height = 15pt, text depth = 15pt] (7.5, 0.2) node {-moe};

\draw[dashed, rounded corners] (1.9, 6.1) rectangle (5.1, 0.9);

\path [->] (4, -1.6) edge (4, -0.4);

\draw (0, -4) rectangle (8, -2);
\draw (4, -3) node {ConveRT};
\path [->] (8.4, -3) edge (9, -3);
\draw [text centered, text width=2.5cm] (10.6, -3) node {Sentence encoding};

\path [->] (4, -5.6) edge (4, -4.4);

\draw (4, -6.4) node {My name is Joseph Schmoe};
\end{tikzpicture}
\caption{Span-ConveRT model architecture. Contextual subword embeddings, computed by ConveRT, are augmented with token features, and fed through a CNN. The outputs of the CNN parameterise a CRF sequence model, defining a distribution over sequence tag labellings, using the \emph{before}, \emph{begin}, \emph{inside}, \emph{after} scheme. Dashed lines denote CNN kernels.}
\label{fig:span_diagram}
\vspace{-1mm}
\end{figure}

\vspace{1.6mm}
\noindent \textbf{Span ConveRT: Final Model.} 
We now describe our model architecture, illustrated in Figure~\ref{fig:span_diagram}. Our approach builds on established sequence tagging models using Conditional Random Fields (CRFs) \cite{ma2016endtoend,lample-etal-2016-neural}. We propose to replace the LSTM part of the model with fixed ConveRT embeddings.\footnote{LSTMs are known to be computationally expensive and require large amounts of resources to obtain any notable success \cite{Pascanu:2013:DTR:3042817.3043083}. By utilizing ConveRT instead, we arrive at a much more lightweight and efficient model.} We take contextualized subword embeddings from ConveRT, giving a sequence of the same length as the subword-tokenized sentence. For sequence tagging, we train a CNN and CRF on top of these fixed subword representations. We concatenate three binary features to the subword representations to emphasize important textual characteristics: (1) whether the token is alphanumeric, (2) numeric, or (3) the start of a new word. In addition, we concatenate the character length of the token as another integer feature. To incorporate the requested slots feature, we concatenate a binary feature representing if the slot is requested to each embedding in the sequence. To contextualize the modified embeddings, we apply a dropout layer followed by a series of 1D convolutions of increasing filter width. 




 Spans are represented using a sequence of \textit{tags}, indicating which members of the subword token sequence are in the span. We use a tag representation similar to the IOB format annotating the span with a sequence of \textit{before}, \textit{begin}, \textit{inside} and \textit{after} tags, see Figure~\ref{fig:span_diagram} for an example.

The distribution of the tag sequence is modeled with a CRF, whose parameters are predicted by a CNN that runs over the contextualized subword embeddings $\mathbf{v}$. At each step $t$, the CNN outputs a $4 \times 4$ matrix of transition scores $\mathbf{W}_t$ and a $4$-dimensional vector of unary potentials $\mathbf{u}_t$. The probability of a predcited tag sequence \(\mathbf{y}\) is then modeled as:
%
{\normalsize
\begin{align*}
    p(\mathbf{y}| \mathbf{v}) \propto  \prod_{t=1}^{T-1}{
        \exp\left(
            \mathbf{W}_t |{y_{t+1},y_{t}}
        \right)}
        \prod_{t=1}^{T}{
        \exp \left( 
            \mathbf{u}_t |{y_{t}}
        \right)
        }
\end{align*}}%
%
\noindent The loss is the negative log-likelihood, equal to minus the sum of the transition scores and unary potentials that correspond to the true tag labels, up to a normalization term. The top scoring tag sequences can be computed efficiently using the Viterbi algorithm.




%% file: 03-experimental.tex
\begin{table}[!t]

\resizebox{0.48\textwidth}{!}{%
\begin{tabular}{r cccccc} \toprule
      & people   & time     & date     & first\_name & last\_name & total \\ \midrule
train & \textbf{2164} (547) & \textbf{2164} (547) & \textbf{1721} (601) &  \textbf{887} (364)    & \textbf{891} (353)    & 8198  \\
dev   & \textbf{983} (244)  & \textbf{853} (276)  & \textbf{802} (300)  &  \textbf{413} (177)    & \textbf{426} (174)    & 3731
\\ \bottomrule
\end{tabular}}
\vspace{-1mm}
\caption{The number of examples for each slot in the \textsc{restaurants-8k} data set. Numbers in brackets show how many examples have the slot requested.}
\label{tab:Restaurants8kDescription}
\vspace{-1mm}
\end{table}

\begin{table*}[!ht]
\def\arraystretch{0.93}
{\small
\begin{tabularx}{\textwidth}{l XXX}
\toprule
\textbf{Hyperparameter}  & \textbf{ConveRT} &\textbf{BERT} & \textbf{Vanilla}   \\ \cmidrule(lr){2-4}
Dimensionality of the input subword embeddings                    & 512           & 768        & 32              \\ 
Size of minibatches during training                               & 64            & 64         & 64              \\ 
The learning rate for the SGD optimizer                           & 0.1          & 0.1       & 0.1             \\ 
Keep probability of elements in the sub-word embedding          & 0.5           & 0.5        & 0.5             \\ 
Keep probability of elements in the sub-word feature embeddings & 0.6           & 0.6        & 0.5             \\ 
The size of the subword-CNN filters                               & (128, 64)     & (128, 64)  & (100, 100, 100) \\ 
Width of the subword CNN filters                                  & (1, 5)        & (1, 5)     & (8, 4, 1)       \\ 
Activation function for subword CNN                               & SiLU         & SiLU      & SiLU    \\      
\bottomrule
\end{tabularx}}%
\centering
\caption{The final hyper-parameters used for different subword representations; \textit{SiLU} refers to the Sigmoid Linear Unit from \newcite{hendrycks2016gelu}.}
\label{tab:hyperparameters}
\vspace{-1.5mm}
\end{table*}

\noindent \textbf{New Evaluation Data Set: \textsc{restaurants-8k}.} 
Data sets for task-oriented dialog systems typically annotate slots with exclusively categorical labels \cite{Budzianowski:2018emnlp}. While some data sets such as SNIPS \cite{coucke2018snips} or ATIS \cite{atis} do contain span annotations, they are built with single-utterance voice commands in mind rather than a natural multi-turn dialog. To fill this gap and enable more work on span extraction for dialog, we introduce a new data set called \textsc{restaurants-8k}. It comprises conversations from a commercial restaurant booking system, and covers 5 slots essential for the booking task: \textit{date}, \textit{time}, \textit{people}, \textit{first name}, \textit{last name}. The data statistics are provided in Table~\ref{tab:Restaurants8kDescription}.\footnote{The data set contains some challenging examples where multiple values are mentioned, or values are mentioned that do not pertain to a slot. For example, in the utterance \textit{``I said 5pm not 6pm''} multiple times are mentioned; in \textit{``I called earlier today''} a date is mentioned that is not the day of the booking. Further, there are noticeable differences compared to previous data sets such as DSTC8 \cite{rastogi2019towards}: e.g., while all slots in other datasets which pertained to integers (e.g. the number of travelers for a coach journey, number of tickets for an event booking) are modeled categorically (i.e. all numbers from 1 to 10 are separate classes), we model the number of people coming for a booking using spans because people often mention this value indirectly. For example \textit{me and my husband}, \textit{3 adults, 4 kids}, \textit{2 couples}.}


\begin{table}[!t]
\def\arraystretch{0.91}
\resizebox{0.47\textwidth}{!}
{%
\begin{tabular}{r ccc}
\toprule
\textbf{Fraction} & \textbf{Span-ConveRT} & \textbf{V-CNN-CRF} & \textbf{Span-BERT} \\ \midrule
1 (8198)        & \textbf{0.96}         & 0.94         & 0.93       \\
$\nicefrac{1}{2}$ (4099)     & \textbf{0.94}         & 0.92         & 0.91       \\
$\nicefrac{1}{4}$ (2049)     & \textbf{0.91}         & 0.89         & 0.88       \\
$\nicefrac{1}{8}$ (1024)     &\textbf{0.89}         & 0.85         & 0.85       \\
$\nicefrac{1}{16}$ (512)     & \textbf{0.81}         & 0.74         & 0.77       \\
$\nicefrac{1}{32}$ (256)     &\textbf{0.64}         & 0.57         & 0.54       \\
$\nicefrac{1}{64}$ (128)     & \textbf{0.58}         & 0.37         & 0.42       \\
$\nicefrac{1}{128}$ (64)    & \textbf{0.41}         & 0.26         & 0.30       \\
\bottomrule
\end{tabular}
}
\caption{Average $F_1$ scores across all slots for \textsc{restaurants-8k} with varying training set fractions. Numbers in brackets represent training set sizes.}
\label{tab:restaurantsResults}
\vspace{-1mm}
\end{table}

\vspace{1.6mm}
\noindent \textbf{DSTC8 Data Sets.}
The Schema-Guided Dialog Dataset (SGDD) \cite{rastogi2019towards} released for \textsc{dstc8} contains span annotations for a subset of slots. We extract span annotated data sets from SGDD in four different domains based on their large variety of slots: (1) \textit{bus and coach booking} (labelled \textit{Buses\_1}), (2) \textit{buying tickets for events} (\textit{Events\_1}), (3) \textit{property viewing} (\textit{Homes\_1}) and \textit{renting cars} (\textit{RentalCars\_1}). A detailed description of the data extraction protocol and the statistics of the data sets, also released with this paper, are available in appendix \ref{s:dstc8_extraction}.

\vspace{1.6mm}
\noindent \textbf{Baseline Models.}
We compare our proposed model with two strong baselines: \textbf{V-CNN-CRF} is a vanilla approach that uses no pretrained model and instead learns sub-word representations from scratch. \textbf{Span-BERT} uses fixed BERT subword representations. All use the same CNN+CRF architecture on top of the subword representations. For each baseline, we conduct hyper-parameter optimization similar to Span-ConveRT: this is done via grid search and evaluation on the development set of \textsc{restaurants-8k}. The final sets of hyper-parameters are provided in Table~\ref{tab:hyperparameters}. Span-BERT relies on BERT-base, with 12 transformer layers and 768-dim embeddings. ConveRT uses 6 transformer layers with 512-dim embeddings, so it is roughly 3 times smaller.

Following prior work \cite{coucke2018snips,rastogi2019towards}, we report the $F_1$ scores for extracting the correct span per user utterance. If the models extract part of the span or a longer span, this is treated as an incorrect span prediction.

\begin{table}[!t]
\def\arraystretch{0.91}
\resizebox{0.47\textwidth}{!}
{%
\begin{tabular}{p{4mm} cccc}
\toprule

       & \textbf{Fraction} & \textbf{Span-ConveRT} & \textbf{V-CNN-CRF} & \textbf{Span-BERT} \\ \midrule
Buses\_1  \\
& 1 (1133)       & \textbf{0.94}         & {0.93}         & 0.93      \\
              & \nicefrac{1}{2} (566)      & \textbf{0.89}         & 0.83         & 0.85      \\
              & \nicefrac{1}{4} (283)     & \textbf{0.84}         & 0.77         & 0.78      \\
              & \nicefrac{1}{8} (141)      & \textbf{0.69}         & 0.68         & 0.70      \\
              & \nicefrac{1}{16} (70)      & \textbf{0.58}         & 0.48         & 0.44      \\ \midrule
Events\_1 \\
& 1 (1498)       & \textbf{0.93}         & {0.92}         & 0.84      \\
              & \nicefrac{1}{2} (749)     & \textbf{0.87}         & 0.82         & 0.80      \\
              & \nicefrac{1}{4} (374)      & \textbf{0.82}         & 0.78         & 0.79      \\
              & \nicefrac{1}{8} (187)      & \textbf{0.70}         & 0.66         & 0.57      \\
              & \nicefrac{1}{16} (93)     & \textbf{0.56}         & 0.50         & 0.44      \\ \midrule
Homes\_1  \\
              & 1 (2064)                    &{ 0.95 }        & 0.92         & \textbf{0.96}      \\
              & \nicefrac{1}{2} (1032)      & \textbf{0.96}         & 0.89         & \textbf{0.96}      \\
              &\nicefrac{1}{4} (516)     & \textbf{0.95}         & 0.85         & \textbf{0.95}      \\
              & \nicefrac{1}{8} (258)       & \textbf{0.93}         & 0.83         & 0.90      \\
              &\nicefrac{1}{16} (129)     & \textbf{0.86}         & 0.67         & 0.76      \\ \midrule
RentalCars\_1 \\
& 1 (874)                                  & \textbf{0.94}         & 0.88         & 0.93      \\
              & \nicefrac{1}{2} (437)      & \textbf{0.93}         & 0.83         & 0.92      \\
              & \nicefrac{1}{4} (218)     & {0.83}         & 0.67         & \textbf{0.86}      \\
              & \nicefrac{1}{8} (109)       & {0.66}         & 0.56         & \textbf{0.79}      \\
              & \nicefrac{1}{16} (54)     & {0.51}         & 0.33         & \textbf{0.61}         \\
              \bottomrule
\end{tabular}}

\hspace{-5mm}
\caption{Average $F_1$ scores on the \textsc{dstc8} single-domain datasets. A full breakdown of results for each individual slot is available in appendix \ref{s:appendixResults}.}
\label{tab:dstc8Results}
\vspace{-1.5mm}
\end{table}

\vspace{1.6mm}
\noindent \textbf{Few-Shot Scenarios.}
For both data sets, we measure performance on smaller sets sampled from the full data. We gradually decrease training sets in size whilst maintaining the same test set: this provides insight on performance in low-data regimes.






%% file: 04-results.tex

The results across all slots are summarized in Table~\ref{tab:restaurantsResults} for \textsc{restaurants-8k}, and in Table~\ref{tab:dstc8Results} for \textsc{dstc8}. First, we note the usefulness of conversational pretraining and transferred representations: Span-ConveRT outperforms the two baselines in all evaluation runs on \textsc{restaurants-8k} and for a large number of runs on \textsc{dstc8}. The gains over V-CNN-CRF directly suggest the importance of transferred pretrained \textit{conversational} representations. Second, we note prominent gains with Span-ConveRT especially in few-shot scenarios with reduced training data: e.g., the gap over V-CNN-CRF widens from 0.02 on the full \textsc{restaurants-8k} training set to 0.15 when using only 64 training examples.\footnote{We also note that Span-BERT outperforms V-CNN-CRF in general for lower-data setups, again indicating the importance of pretraining especially for such setups. However, conversational pretraining with ConveRT still yields the strongest overall performance.} Similar trends are observed on all four \textsc{dstc8} subsets. Again, this indicates 1) importance of pretraining in general for low-data regimes, and 2) that general-purpose conversational knowledge coded in ConveRT can indeed boost dialog modeling in such low-data regimes. If sufficient domain-specific data is available (e.g., see the results of V-CNN-CRF with full data), learning domain-specialized representations from scratch can lead to strong performance, but using transferred conversational representations seems to be widely useful and robust.

We also report gains over Span-BERT for a large number of runs, and weaker performance of Span-BERT in comparison to V-CNN-CRF in many higher-data setups (see Table~\ref{tab:restaurantsResults}). These results indicate that for conversational end-applications such as slot-filling, pretraining on a \textit{conversational} task (such as response selection) is more beneficial than standard language modeling-based pretraining. Our hypothesis is that both the vanilla baseline and ConveRT leverage some ``domain adaptation'': ConveRT is trained on rich conversational data, while the baseline representations are learned directly on the training data. BERT, on the other hand, is not trained on conversational data directly and usually relies on much longer passages of text. This might not make the BERT representations suitable for conversational tasks such as span extraction. Similar findings, where ConveRT-based conversational representations outperform BERT-based baselines (even with full fine-tuning), have recently been established in other dialog tasks such as intent detection \cite{Henderson:2019convert,Casanueva:2020ws,Bunk:2020arxiv}. In general, our findings also call for investing more effort in investigating different pretraining strategies that are better aligned to target tasks \cite{Mehri:2019acl,Henderson:2019convert, Humeau:2019arxiv}.

\vspace{1.6mm}
\noindent \textbf{Error Analysis.} To better understand the performance of Span-ConveRT on the \textsc{restaurants-8k} data set, we also conducted a manual error analysis, comparing it with the best performing baseline model, V-CNN-CRF. In Appendix~\ref{s:appendixQuantitativeResults} we lay out the types of errors that occur in a generic span extraction task and investigate the distribution of these types of errors across slots and models. We show that when trained in the high-data setting the distribution is similar between the two models, suggesting that gains from Span-ConveRT are across all types of error. We also show that the distribution varies more in the low-data setting and discuss how that might impact their comparative performance in practice. Additionally, in Appendix~\ref{s:appendixQualitativeResults} we provide a qualitative analysis on the errors the two models make for the slot \textit{first name}. We show that the baseline model has a far greater tendency to wrongly identify generic out-of-vocabulary words as names.

%% file: 06-conclusion.tex
We have introduced \textit{Span-ConveRT}, a light-weight model for dialog slot-filling that approaches the problem as a turn-based span extraction task. The formulation allows the model to effectively leverage representations available from large-scale conversational pretraining. We have shown that, due to pretrained representations, Span-ConveRT is especially useful in few-shot learning setups on small data sets. We have also introduced \textsc{restaurants-8k}, a new challenging data set that will hopefully encourage further work on span extraction for dialogue. In future work, we plan to experiment with multi-domain span extraction architectures.




%% file: 07-appendix.tex
\appendix
\twocolumn

\section{DSTC8 Datasets: Data Extraction and Statistics}\label{s:dstc8_extraction}
As discussed in \S\ref{s:exp}, we extract span annotated data sets from the Schema Guided Dialog Dataset (SGDD) in four different domains. SGDD is a multi-domain data set with each domain consisting of several sub-domains. As the data set has been built for transfer learning from one domain to another, many sub-domains only exist in either the training or development data sets. We are interested in single-domain dialog, and therefore chose datasets from four different domains of the original dataset: (1) \textit{bus and coach booking}, (2) \textit{buying tickets for events}, (3) \textit{property viewing} and \textit{renting cars}. We select these domains due to their high number of conversations and their large variety of slots (e.g. \textit{area of city to view an apartment}, \textit{type of event to attend}, \textit{time/date of coach to book}). For each of these domains, we chose their first sub-domain\footnote{ We refer to them by their corresponding ID in the original data set: \textit{Buses\_1}, \textit{Events\_1}, \textit{Homes\_1}, \textit{RentalCars\_1}}, and took all turns from conversations that stay within this sub-domain. For the requested slots feature, we check for when the system action of the turn prior contains a \mytexttt{REQUEST} action. The training and development split is kept the same for all extracted turns. Table~\ref{tab:dstc8Description} shows the resulting data set sizes for each sub-domain. We are releasing these filtered single-domain data sets, along with the code to create them from the original SGDD data.

\begin{table*}[!t]
\begin{center}
{\small
\renewcommand*{\arraystretch}{1.8}
\begin{tabular}{r c c p{60mm}}

\toprule
   \textbf{Sub-domain} & \textbf{Train Size} & \textbf{Dev Size} & \textbf{Slots}                                                                                                                     \\ \midrule
\textbf{Buses\_1}      & 1133                & 377               & 
from\_location (169/54), leaving\_date (165/57),
to\_location (166/52)                  \\ 
\textbf{Events\_1}     & 1498                & 521               & 
city\_of\_event (253/82), date (151/33), 
subcategory (56/26)                               \\ 
\textbf{Homes\_1}      & 2064                & 587               & area (288/86), visit\_date (237/62)                                                                                                  \\ 
\textbf{RentalCars\_1} & 874                 & 328               & 
dropoff\_date (112/42), pickup\_city (116/48),  pickup\_date (120/43), pickup\_time (119/43)\\
\bottomrule
\end{tabular}
}
\end{center}
\caption{Statistics of the used data sets extracted from the DSTC8 schema-guided dialog dataset. We also report the number of examples in the train and development sets for each slot in parentheses.}
\label{tab:dstc8Description}
\vspace{-5mm}
\end{table*}

\clearpage
\onecolumn
\section{Experimental Results on \textsc{Restaurants-8k} and \textsc{dstc8}: $F_1$ Scores for Each Slot}
\label{s:appendixResults}

\begin{table}[h]
\centering
\resizebox{0.65\textwidth}{!}{%
\begin{tabular}{l|cccc}
\textbf{Slot}        & \textbf{Fraction} &  \textbf{Span-ConveRT}  & \textbf{V-CNN-CRF}   & \textbf{Span-BERT} \\ \hline
date        & 1        & 0.95 & 0.95 & 0.93 \\
            & 1/2      & 0.95 & 0.94 & 0.91 \\
            & 1/4      & 0.92 & 0.93 & 0.88 \\
            & 1/8      & 0.90 & 0.87 & 0.86 \\
            & 1/16     & 0.85 & 0.83 & 0.78 \\
            & 1/32     & 0.78 & 0.73 & 0.69 \\
            & 1/64     & 0.73 & 0.54 & 0.61 \\
            & 1/128    & 0.61 & 0.40 & 0.44 \\ \hline
first\_name & 1        & 0.96 & 0.92 & 0.95 \\
            & 1/2      & 0.94 & 0.92 & 0.91 \\
            & 1/4      & 0.92 & 0.88 & 0.90 \\
            & 1/8      & 0.89 & 0.85 & 0.87 \\
            & 1/16     & 0.81 & 0.64 & 0.73 \\
            & 1/32     & 0.57 & 0.36 & 0.28 \\
            & 1/64     & 0.49 & 0.21 & 0.23 \\
            & 1/128    & 0.22 & 0.06 & 0.04 \\ \hline
last\_name  & 1        & 0.96 & 0.91 & 0.93 \\
            & 1/2      & 0.94 & 0.88 & 0.94 \\
            & 1/4      & 0.90 & 0.83 & 0.89 \\
            & 1/8      & 0.88 & 0.78 & 0.87 \\
            & 1/16     & 0.80 & 0.62 & 0.77 \\
            & 1/32     & 0.49 & 0.43 & 0.40 \\
            & 1/64     & 0.36 & 0.04 & 0.16 \\
            & 1/128    & 0.13 & 0.06 & 0.02 \\ \hline
people      & 1        & 0.97 & 0.94 & 0.93 \\
            & 1/2      & 0.94 & 0.93 & 0.90 \\
            & 1/4      & 0.92 & 0.91 & 0.86 \\
            & 1/8      & 0.88 & 0.87 & 0.84 \\
            & 1/16     & 0.83 & 0.78 & 0.78 \\
            & 1/32     & 0.73 & 0.63 & 0.63 \\
            & 1/64     & 0.69 & 0.50 & 0.54 \\
            & 1/128    & 0.61 & 0.41 & 0.47 \\ \hline
time        & 1        & 0.95 & 0.95 & 0.93 \\
            & 1/2      & 0.94 & 0.93 & 0.90 \\
            & 1/4      & 0.90 & 0.91 & 0.86 \\
            & 1/8      & 0.87 & 0.87 & 0.84 \\
            & 1/16     & 0.77 & 0.78 & 0.78 \\
            & 1/32     & 0.61 & 0.63 & 0.63 \\
            & 1/64     & 0.60 & 0.50 & 0.54 \\
            & 1/128    & 0.46 & 0.41 & 0.47
\end{tabular}}
\label{tab:allRestaurants8kResults}
\caption{F1 scores for each slot in the Restaurants8k datastet.}
\end{table}

\begin{table}[t]
\centering
\resizebox{0.7\textwidth}{!}{%
    \begin{tabular}{l|ccccc}
    \textbf{Dataset} & \textbf{Slot}            & \textbf{Fraction} & \textbf{ConveRT Reps} & \textbf{Vanilla Reps} & \textbf{BERT Reps} \\ \hline
    Buses\_1          & from\_location            & 1                            & 0.94                             & 0.93                             & 0.89                           \\
    ~                           & ~                                   & 1/2                          & 0.90                             & 0.77                             & 0.77                           \\
    ~                           & ~                                   & 1/4                          & 0.86                             & 0.73                             & 0.73                           \\
    ~                           & ~                                   & 1/8                          & 0.63                             & 0.55                             & 0.53                           \\
    ~                           & ~                                   & 1/16                         & 0.48                             & 0.49                             & 0.29                           \\
    ~                           & leaving\_date             & 1                            & 0.98                             & 0.94                             & 0.96                           \\
    ~                           & ~                                   & 1/2                          & 0.99                             & 0.91                             & 0.97                           \\
    ~                           & ~                                   & 1/4                          & 0.96                             & 0.88                             & 0.91                           \\
    ~                           & ~                                   & 1/8                          & 0.86                             & 0.80                             & 0.91                           \\
    ~                           & ~                                   & 1/16                         & 0.77                             & 0.57                             & 0.59                           \\
    ~                           & to\_location              & 1                            & 0.88                             & 0.92                             & 0.94                           \\
    ~                           & ~                                   & 1/2                          & 0.77                             & 0.81                             & 0.81                           \\
    ~                           & ~                                   & 1/4                          & 0.69                             & 0.69                             & 0.69                           \\
    ~                           & ~                                   & 1/8                          & 0.58                             & 0.69                             & 0.65                           \\
    ~                           & ~                                   & 1/16                         & 0.50                             & 0.37                             & 0.46                           \\ \hline
    Events\_1         & city\_of\_event & 1                            & 0.93                             & 0.95                             & 0.94                           \\
    ~                           & ~                                   & 1/2                          & 0.93                             & 0.93                             & 0.91                           \\
    ~                           & ~                                   & 1/4                          & 0.88                             & 0.78                             & 0.88                           \\
    ~                           & ~                                   & 1/8                          & 0.83                             & 0.79                             & 0.73                           \\
    ~                           & ~                                   & 1/16                         & 0.73                             & 0.73                             & 0.64                           \\
    ~                           & date                                & 1                            & 0.91                             & 0.88                             & 0.88                           \\
    ~                           & ~                                   & 1/2                          & 0.89                             & 0.91                             & 0.91                           \\
    ~                           & ~                                   & 1/4                          & 0.68                             & 0.78                             & 0.81                           \\
    ~                           & ~                                   & 1/8                          & 0.72                             & 0.78                             & 0.67                           \\
    ~                           & ~                                   & 1/16                         & 0.66                             & 0.52                             & 0.68                           \\
    ~                           & subcategory                         & 1                            & 0.94                             & 0.94                             & 0.71                           \\
    ~                           & ~                                   & 1/2                          & 0.79                             & 0.62                             & 0.59                           \\
    ~                           & ~                                   & 1/4                          & 0.81                             & 0.77                             & 0.67                           \\
    ~                           & ~                                   & 1/8                          & 0.55                             & 0.41                             & 0.32                           \\
    ~                           & ~                                   & 1/16                         & 0.29                             & 0.25                             & 0.00                           \\ \hline
    Homes\_1          & area                                & 1                            & 0.96                             & 0.95                             & 0.94                           \\
    ~                           & ~                                   & 1/2                          & 0.94                             & 0.88                             & 0.95                           \\
    ~                           & ~                                   & 1/4                          & 0.93                             & 0.87                             & 0.94                           \\
    ~                           & ~                                   & 1/8                          & 0.91                             & 0.76                             & 0.85                           \\
    ~                           & ~                                   & 1/16                         & 0.80                             & 0.67                             & 0.69                           \\
    ~                           & visit\_date               & 1                            & 0.94                             & 0.89                             & 0.98                           \\
    ~                           & ~                                   & 1/2                          & 0.98                             & 0.89                             & 0.97                           \\
    ~                           & ~                                   & 1/4                          & 0.98                             & 0.84                             & 0.96                           \\
    ~                           & ~                                   & 1/8                          & 0.96                             & 0.90                             & 0.94                           \\
    ~                           & ~                                   & 1/16                         & 0.93                             & 0.67                             & 0.83                           \\ \hline
    RentalCars\_1     & dropoff\_date             & 1                            & 0.91                             & 0.89                             & 0.92                           \\
    ~                           & ~                                   & 1/2                          & 0.93                             & 0.86                             & 0.90                           \\
    ~                           & ~                                   & 1/4                          & 0.71                             & 0.55                             & 0.74                           \\
    ~                           & ~                                   & 1/8                          & 0.56                             & 0.74                             & 0.41                           \\
    ~                           & ~                                   & 1/16                         & 0.55                             & 0.60                             & 0.48                           \\
    ~                           & pickup\_city              & 1                            & 0.98                             & 0.84                             & 0.93                           \\
    ~                           & ~                                   & 1/2                          & 0.98                             & 0.74                             & 0.89                           \\
    ~                           & ~                                   & 1/4                          & 0.94                             & 0.60                             & 0.87                           \\
    ~                           & ~                                   & 1/8                          & 0.73                             & 0.51                             & 0.65                           \\
    ~                           & ~                                   & 1/16                         & 0.44                             & 0.17                             & 0.38                           \\
    ~                           & pickup\_date              & 1                            & 0.90                             & 0.85                             & 0.88                           \\
    ~                           & ~                                   & 1/2                          & 0.85                             & 0.76                             & 0.81                           \\
    ~                           & ~                                   & 1/4                          & 0.75                             & 0.71                             & 0.79                           \\
    ~                           & ~                                   & 1/8                          & 0.65                             & 0.55                             & 0.68                           \\
    ~                           & ~                                   & 1/16                         & 0.40                             & 0.29                             & 0.39                           \\
    ~                           & pickup\_time              & 1                            & 0.98                             & 0.95                             & 0.98                           \\
    ~                           & ~                                   & 1/2                          & 0.97                             & 0.96                             & 0.91                           \\
    ~                           & ~                                   & 1/4                          & 0.92                             & 0.82                             & 0.85                           \\
    ~                           & ~                                   & 1/8                          & 0.71                             & 0.41                             & 0.73                           \\
    \end{tabular}
}%
\caption{F1 scores for all of the slots in the DSTC8 single-domain experiments.}
\label{tab:alldstc8Results}
\end{table}
\fi

\clearpage
\twocolumn
\section{Quantitative Error Analysis of Span-ConveRT and V-CNN-CRF on \textsc{restaurants-8k}}
\label{s:appendixQuantitativeResults}

We divide the errors into four categories:
\newenvironment{packed_enum}{
\begin{enumerate}
  \setlength{\itemsep}{1pt}
  \setlength{\parskip}{0pt}
  \setlength{\parsep}{0pt}
}{\end{enumerate}}
\begin{packed_enum}
  \item The model predicted no span when there was a span present.
 \item The model predicted a span when no span was present.
 \item The model predicted a span which does not overlap the label span.
 \item The model predicted a span which overlaps label span.
\end{packed_enum}

\begin{figure*}[b!]
    \includegraphics[width=0.99\textwidth]{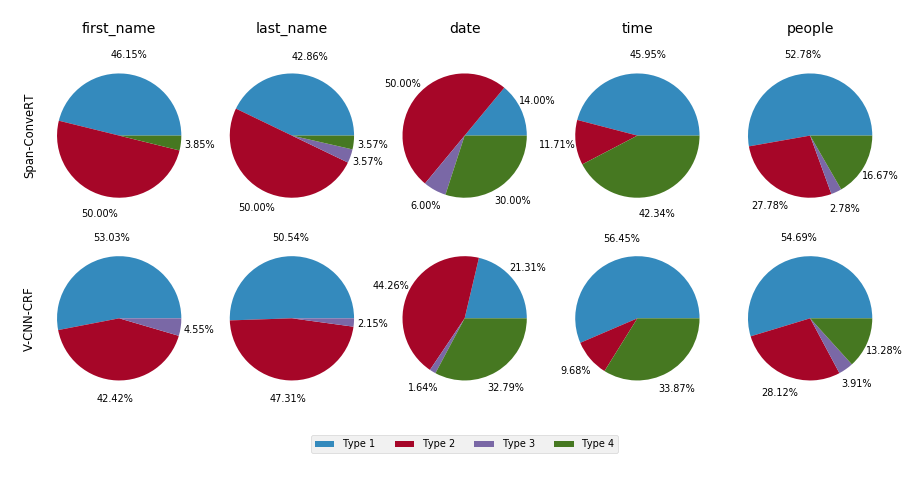} 
    \centering
    \caption{Breakdown of errors made on the test set of \textsc{restaurants-8k} after training on the entire train set.}
    \label{fig:error_full_data}
    \vspace{-1mm}
\end{figure*}

When training on the full training set (Figure~\ref{fig:error_full_data}), there is little difference in error breakdown between Span-ConveRT and V-CNN-CRF. This suggests the behavior of these models is similar when trained in a high-data setting, but improvements made by Span-ConveRT are on all fronts. 

\begin{figure*}[b!]
    \includegraphics[width=0.99\textwidth]{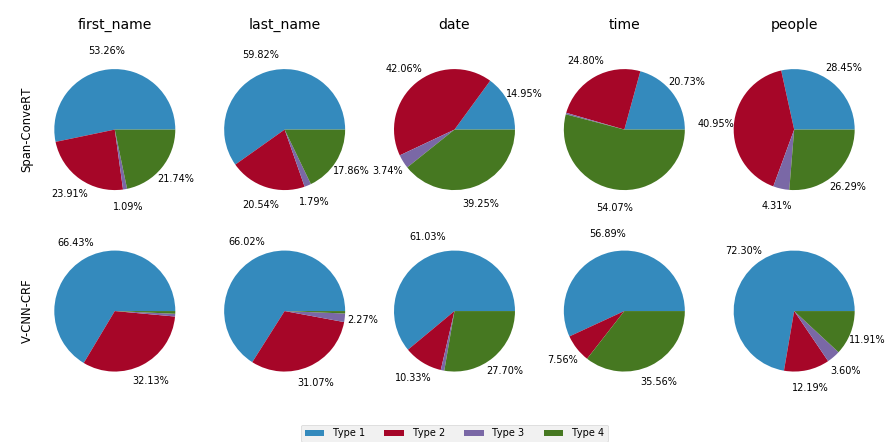} 
    \centering
    \caption{Breakdown of errors made on the test set of \textsc{restaurants-8k} after training on a 16th of the train set.}
    \label{fig:error_16th_data}
    \vspace{-1mm}
\end{figure*}

When trained on a 16th of the dataset (Figure~\ref{fig:error_16th_data}), the difference between the models becomes more pronounced. Most notably, the Span-ConveRT model produces a greater proportion of type 4 errors compared to the V-CNN-CRF model on every slot. This suggests that the errors Span-ConveRT makes, although not precisely correct with its span prediction, are more likely to yield a span that could parse to a correct value. For example, consider the sentence ``a table for 8pm this evening''. The correct span for the slot \textit{time} is "8pm", but if a model erroneously predicts ``8pm this evening'' (a span which overlaps the label span) it will still parse to the same time as the label span.

\clearpage
\twocolumn
\section{Qualitative Error Analysis of Span-ConveRT and V-CNN-CRF on \textsc{restaurants-8k}}
\label{s:appendixQualitativeResults}
As an accompaniment to the quantitative results, we provide a brief qualitative analysis of errors in the best performing models. Considering only the \textit{first name} slot, we collect the errors made on the test set that are exclusive to each model. That left 10 errors for Span-ConveRT and 50 for V-CNN-CRF. Along with our analysis based on the full set of 60 errors, we provide a random sample of 5 errors from each model in Tables~\ref{tab:qual_err_spanconvert} and \ref{tab:qual_err_vcnncrf}.

\begin{table*}[h]
\begin{center}
{\small
\renewcommand*{\arraystretch}{1.8}
\begin{tabular}{r l}
\toprule
\textbf{Probability} & \textbf{Text/Spans}\\
\midrule
N/A & \textcolor{orange}{Wen} Books, for 7:15PM, I made a reservation yesterday for a party of 8\\
0.4447 & \textcolor{red}{Saul}\\
0.9685 & \textcolor{red}{Adragna}\\
0.9247 & last name \textcolor{red}{Prader}\\
0.9553 & \textcolor{red}{Verjan}\\
\bottomrule
\end{tabular}
}
\end{center}
\caption{Random sample of errors exclusively made by \textbf{Span-ConveRT} for the slot \textit{first name}. Red text denotes incorrectly predicted spans and orange denotes true spans that were not predicted.}
\label{tab:qual_err_spanconvert}
\end{table*}

\begin{table*}[h]
\begin{center}
{\small
\renewcommand*{\arraystretch}{1.8}
\begin{tabular}{r l}
\toprule
\textbf{Probability} & \textbf{Text/Spans}\\
\midrule
0.8872 & \textcolor{red}{bloody} useless\\
0.3939 & What is their \textcolor{red}{web} URL?\\
0.3319 & ok are you guys \textcolor{red}{animal} friendly\\
0.8604 & My 7 friends and I can \textcolor{red}{spread} ourselves over two tables if necessary\\
N/A & \textcolor{orange}{Gertrudis} Hayslett\\
\bottomrule
\end{tabular}
}
\end{center}
\caption{Random sample of errors exclusively made by \textbf{V-CNN-CRF} for the slot \textit{first name}. Red text denotes incorrectly predicted spans and orange denotes true spans that were not predicted.}
\label{tab:qual_err_vcnncrf}
\end{table*}

A large portion of the errors exclusively made by V-CNN-CRF were predictions of spans where no name was mentioned. Many words that are not standard to the domain of restaurant booking were, often confidently, wrongly predicted as names. For example, in Table~\ref{tab:qual_err_vcnncrf} we show that the words ``bloody'', ``web'', ``animal'' and ``spread'' were all predicted as first names by the baseline model. Employing transferred conversational representations evidently lessens the likelihood of these forms of errors occurring. Also included in the table is an example where the baseline model fails to recognize a name which, when corroborated with similar occurrences in the wider set of errors, suggests that it is less likely to predict spans for out-of-vocabulary names than Span-ConveRT.

As well as backing up the conclusions formed by our numerical results, we were also interested in what ways using pretrained representations might hinder performance. With only 10 errors exclusively made by Span-ConveRT it was not possible to form any sweeping conclusions but a handful of errors suggest that the model might employ its background knowledge to reject unfamiliar first names or accept familiar ones in spite of the sentence structure suggesting otherwise. For example, in the first row of Table~\ref{tab:qual_err_spanconvert} we find that the model rejects the name ``Wen'' despite it being part of a fairly common exchange for this domain and in a natural place for a first name. The other examples demonstrate that the model can sometimes predict last names as first names and in spite of contextual cues suggesting otherwise, can do so over-confidently.
